\documentclass[10pt, conference]{IEEEtran}
\ifCLASSINFOpdf
\else
\fi

\usepackage{cite}
\usepackage{amsmath,amssymb,amsfonts}
\usepackage{algorithmic}
\usepackage{graphicx}
\usepackage{textcomp}
\usepackage{xcolor}
\usepackage{url}
\newcommand\EatDot[1]{}
\usepackage[pagebackref=true,colorlinks,bookmarks=false]{hyperref}
\definecolor{OliveGreen}{rgb}{0,0.6,0}
\definecolor{SoftRed}{rgb}{1,0.2,0.2}

\hypersetup{
     colorlinks   = true,
     citecolor    = OliveGreen,
     linkcolor    = SoftRed
}
\usepackage[hyperpageref]{backref}
\usepackage{tabularx}
\usepackage{booktabs}
\usepackage{multirow}
\usepackage{flushend}
\usepackage{adjustbox}
\usepackage{subcaption}

\begin{document}
%
\title{Indiscapes: Instance Segmentation Networks for Layout Parsing of Historical Indic Manuscripts}




%
\author{\IEEEauthorblockN{Abhishek Prusty,
Sowmya Aitha,
Abhishek Trivedi, 
Ravi Kiran Sarvadevabhatla}
\IEEEauthorblockA{Centre for Visual Information Technology (CVIT)\\
International Institute of Information Technology, Hyderabad (IIIT-H) \\
Gachibowli, Hyderabad 500032, INDIA. \\  \texttt{\small \string{abhishek.prusty@students.,sowmya.aitha@research.,abhishek.trivedi@research.,ravi.kiran@\string}iiit.ac.in}}
}


\maketitle

\begin{abstract}
Historical palm-leaf manuscript and early paper documents from Indian subcontinent form an important part of the world's literary and cultural heritage. Despite their importance, large-scale annotated Indic manuscript image datasets do not exist. To address this deficiency, we introduce Indiscapes, the first ever dataset with multi-regional layout annotations for historical Indic manuscripts. To address the challenge of large diversity in scripts and presence of dense, irregular layout elements (e.g. text lines, pictures, multiple documents per image), we adapt a Fully Convolutional Deep Neural Network architecture for fully automatic, instance-level spatial layout parsing of manuscript images. We demonstrate the effectiveness of proposed architecture on images from the Indiscapes dataset. For annotation flexibility and keeping the non-technical nature of domain experts in mind, we also contribute a custom, web-based GUI annotation tool and a dashboard-style analytics portal. Overall, our contributions set the stage for enabling downstream applications such as OCR and word-spotting in historical Indic manuscripts at scale. 
\end{abstract}


\section{Introduction}

The collection and analysis of historical document images is a key component in the preservation of culture and heritage. Given its importance, a number of active research efforts exist across the world~\cite{reul2017case,springmann2017ocr,simistira2016diva,PappoToledano2018AdoptiveTA,kesiman2016amadi_lontarset,chen2015page}. In this paper, we focus on palm-leaf and early paper documents from the Indian sub-continent. In contrast with modern or recent era documents, such manuscripts are considerably more fragile, prone to degradation from elements of nature and tend to have a short shelf life~\cite{sahoo2016selective,rachman2018palm,kumar2009traditional}. More worryingly, the domain experts who can decipher such content are small in number and dwindling. Therefore, it is essential to access the content within these documents before it is lost forever. 

Surprisingly, no large-scale annotated Indic manuscript image datasets exist for the benefit of researchers in the community. In this paper, we take a significant step to address this gap by creating such a dataset. Given the large diversity in language, script and non-textual regional elements in these manuscripts, spatial layout parsing is crucial in enabling downstream applications such as OCR, word-spotting, style-and-content based retrieval and clustering. For this reason, we first tackle the problem of creating a diverse, annotated \textit{spatial layout} dataset. This has the immediate advantage of bypassing the hurdle of language and script familiarity for annotators since layout annotation does not require any special expertise unlike text annotation.

\begin{figure*}[!tbp]
\includegraphics[width=\textwidth]{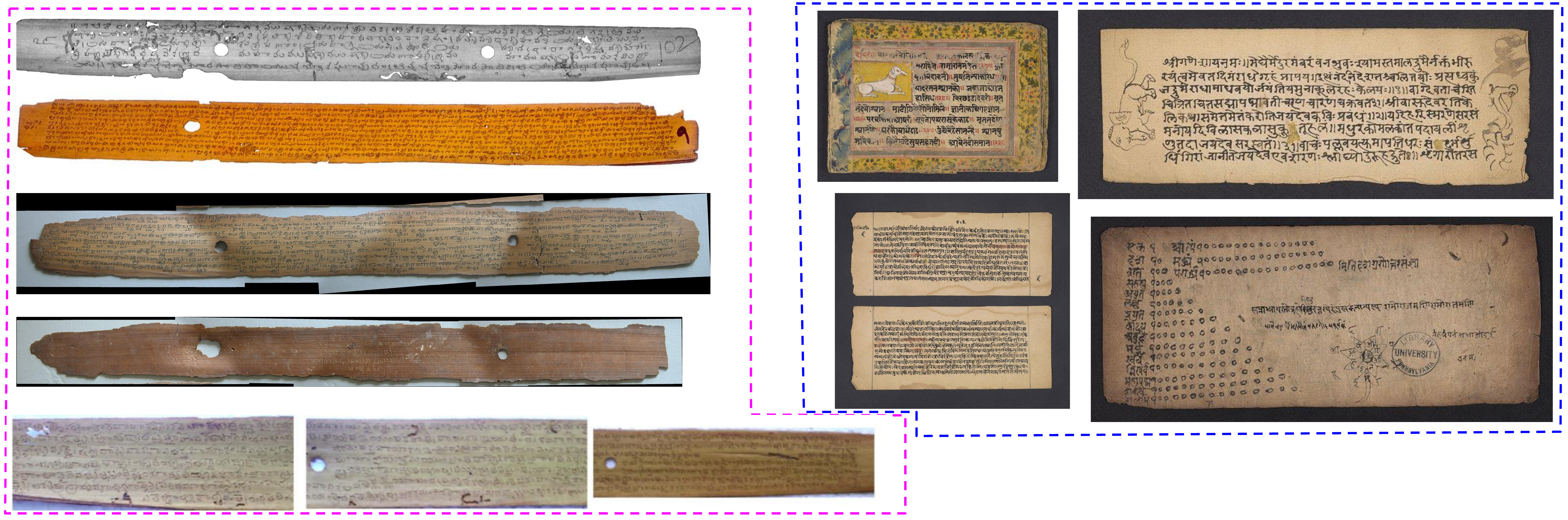}
\caption{The five images on the left, enclosed by pink dotted line, are from the \textsc{Bhoomi} palm leaf manuscript collection while the remaining images (enclosed by blue dotted line) are from the 'Penn-in-Hand' collection (refer to Section \ref{sec:dataset}). Note the inter-collection differences, closely spaced and unevenly written text lines, presence of various non-textual layout regions (pictures, holes, library stamps), physical degradation and presence of multiple manuscripts per image. All of these factors pose great challenges for annotation and machine-based parsing.}
\label{fig:indiscapes-images}
\end{figure*}

In general, manuscripts from Indian subcontinent pose many unique challenges (Figure \ref{fig:indiscapes-images}). To begin with, the documents exhibit a large multiplicity of languages. This is further magnified by variations in intra-language script systems. Along with text, manuscripts may contain pictures, tables, non-pictorial decorative elements in non-standard layouts. A unique aspect of Indic and South-East Asian manuscripts is the frequent presence of holes punched in the document for the purpose of binding~\cite{kumar2009traditional,valy2017new,sahoo2016selective}. These holes cause unnatural gaps within text lines. The physical dimensions of the manuscripts are typically smaller compared to other historical documents, resulting in a dense content layout. Sometimes, multiple manuscript pages are present in a single image. Moreover, imaging-related factors such as varying scan quality play a role as well. Given all of these challenges, it is important to develop robust and scalable approaches for the problem of layout parsing. In addition, given the typical non-technical nature of domain experts who study manuscripts, it is also important to develop easy-to-use graphical interfaces for annotation, post-annotation visualization and analytics. 

We make the following contributions:
\begin{itemize}
    \item We introduce Indiscapes, the first ever historical Indic manuscript dataset with detailed spatial layout annotations (Section \ref{sec:dataset}).
    \item We adapt a deep neural network architecture for instance-level spatial layout parsing of historical manuscript images (Section \ref{sec:layout-parsing-network}).
    \item We also introduce a lightweight web-based GUI for annotation and dashboard-style analytics keeping in mind the non-technical domain experts and the unique layout-level challenges of Indic manuscripts (Section \ref{subsec:annot-tool}).
\end{itemize}

\section{Related Work}

A number of research groups have invested significant efforts in the creation and maintenance of annotated, publicly available historical manuscript image datasets~\cite{sanchez2014handwritten,rath2007word,reul2017case,springmann2017ocr,simistira2016diva,PappoToledano2018AdoptiveTA,kassis2017vmlhd}. Other collections contain character-level and word-level spatial annotations for South-East Asian palm-leaf manuscripts~\cite{valy2017new, kesiman2016amadi_lontarset,suryani2017handwritten}. In these latter set of works, annotations for lines are obtained by considering the polygonal region formed by union of character bounding boxes as a line. While studies on Indic palm-leaf and paper-based manuscripts exist, these are typically conducted on small and often, private collections of documents~\cite{clausner2017icdar2017,Savitha_2018,abeysinghe2018use,sastry20173d,Panyam:2016:MPL:3043545.3064264,shi2004digital,malayalamPalm}. No publicly available large-scale, annotated dataset of historical Indic manuscripts exists to the best of our knowledge. In contrast with existing  collections, our proposed dataset contains a much larger diversity in terms of document type (palm-leaf and early paper), scripts and annotated layout elements (see Tables \ref{tab:dataset-region-stats},\ref{tab:dataset-script-stats}). An additional level of complexity arises from the presence of multiple manuscript pages within a single image (see Fig. \ref{fig:indiscapes-images}).

A number of contributions can also be found for the task of historical document layout parsing~\cite{wick2018fully,Wei2015,bukhari2012layout,chen2017convolutional}. Wei et al.~\cite{Wei2015} explore the effect of using a hybrid feature selection method while using autoencoders for semantic segmentation in five historical English and Medieval European manuscript datasets. Chen et al.~\cite{chen2017convolutional} explore the use of Fully Convolutional Networks (FCN) for the same datasets. Barakat et al.~\cite{barakat2018text} propose a FCN for segmenting closely spaced, arbitrarily oriented text lines from an Arabic manuscript dataset. The mentioned approaches, coupled with efforts to conduct competitions on various aspects of historical document layout analysis have aided progress in this area~\cite{DBLP:conf/icfhr/KesimanVBPSHVCO18,DBLP:conf/icdar/2017hip,DBLP:conf/icdar/2015hip}. 
A variety of layout parsing approaches, including those employing the modern paradigm of deep learning, have been proposed for Indic~\cite{sastry20173d,shi2004digital,Sabeenian:2017:CHT:3133793.3133804,malayalamPalm} and South-East Asian~\cite{bukhari2012layout,kesiman2018benchmarking,suryani2017handwritten,Valy2018Khmer,Paulus2018} palm-leaf and paper manuscript images. However, existing approaches typically employ brittle hand-crafted features or demonstrate performance on datasets which are limited in terms of layout diversity. Similar to many recent works, we employ Fully Convolutional Networks in our approach. However, a crucial distinction lies in our formulation of layout parsing as an \textit{instance} segmentation problem, rather than just a \textit{semantic} segmentation problem. This avoids the problem of closely spaced layout regions (e.g. lines) being perceived as contiguous blobs.

The ready availability of annotation and analysis tools has facilitated progress in creation and analysis of historical document manuscripts~\cite{doermann2010gedi,garz2016creating,clausner2011aletheia}. The tool we propose in the paper contains many of the features found in existing annotation systems. However, some of these systems are primarily oriented towards single-user, offline annotation and do not enable a unified management of annotation process and monitoring of annotator performance. In contrast, our web-based system addresses these aspects and provides additional capabilities. Many of the additional features in our system are tailored for annotation and examining annotation analytics for documents with dense and irregular layout elements, especially those found in Indic manuscripts. In this respect, our annotation system is closer to the recent trend of collaborative, cloud/web-based annotation systems and services~\cite{webaletheia,wursch2016divaservices,gatos2014ground}.

\section{Indiscapes: The Indic manuscript dataset}
\label{sec:dataset}

\begin{table*}[!t]

\resizebox{\textwidth}{!}
{
 \centering 
 \begin{tabular}{cccccccccc}
 \toprule 
                & Character Line Segment  & Character Component  &  Hole & Page Boundary & Library Marker & Decorator & Picture & Physical Degradation
                & Boundary Line\\
                & (CLS)  & (CC)  & (H) & (PB) & (LM) & (D) & (P) & (PD) & (BL) \\
 \midrule   
 \textsc{PIH}  & $2401$ & $494$ & $-$ & $256$ & $32$ & $59$ & $94$ & $34$ & $395$ \\
 \textsc{Bhoomi}      & $2440$ & $210$ & $565$ & $316$ & $133$ & $-$ & $-$ & $2078$& $-$\\
 \midrule   
 Combined      & $4841$ & $704$ & $565$ & $572$ & $165$ & $59$ & $94$ & $2112$ & $395$\\
 \bottomrule
 \end{tabular}
 }
\captionof{table}{Counts for various annotated region types in \textsc{Indiscapes} dataset. The abbreviations used for region types are given below each region type.}
\label{tab:dataset-region-stats} 
\end{table*}

\begin{table}[!ht]
\centering
\resizebox{0.44\textwidth}{!}
{%
  
 \begin{tabular}{cccc|c}
 \toprule 
                & Train  &  Validation & Test  & Total \\
\midrule   
    \textsc{PIH}         & $116$        & $28$        & $49$           & $193$           \\
    \textsc{Bhoomi}      & $236$        & $59$         & $20$           & $315$          \\
    \midrule   
    Total      & $352$       & $87$         & $69$           & $508$            \\
 \bottomrule
 \end{tabular}
 }
\captionof{table}{Dataset splits used for  learning and inference.}
\label{table:document-counts} 
\end{table}

The Indic manuscript document images in our dataset are obtained from two sources. The first source is the publicly available Indic manuscript collection from University of Pennsylvania's Rare Book and Manuscript Library~\cite{penninhand}, also referred to as Penn-in-Hand (\textsc{PIH}). From the $2{,}880$ Indic manuscript book-sets\footnote{A book-set is a sequence of manuscript images.}, we carefully curated $193$ manuscript images for annotation. Our curated selection aims to maximize the diversity of the dataset in terms of various attributes such as the extent of document degradation, script language, presence of non-textual elements (e.g. pictures, tables) and number of lines. Some images contain multiple manuscript pages stacked vertically or horizontally (see bottom-left image in Figure \ref{fig:indiscapes-images}). The second source for manuscript images in our dataset is \textsc{Bhoomi}, an assorted collection of $315$ images sourced from multiple Oriental Research Institutes and libraries across India. As with the first collection, we chose a subset intended to maximize the overall diversity of the dataset. However, this latter set of images are characterized by a relatively inferior document quality, presence of multiple languages and from a layout point of view, predominantly contain long, closely and irregularly spaced text lines, binding holes and degradations (Figure \ref{fig:indiscapes-images}). Though some document images contain multiple manuscripts, we do not attempt to split the image into multiple pages. While this poses a challenge for annotation and automatic image parsing, retaining such images in the dataset eliminates manual/semi-automatic intervention. As our results show, our approach can successfully handle such multi-page documents, thereby making it truly an end-to-end system. 

Overall, our dataset contains $508$ annotated Indic manuscripts. Some salient aspects of the dataset can be viewed in Table ~\ref{tab:dataset-region-stats} and a pictorial illustration of layout regions can be viewed in Figure \ref{fig:results}. Note that multiple regions can overlap, unlike existing historical document datasets which typically contain disjoint region annotations. 

For the rest of the section, we discuss the challenges associated with annotating Indic manuscripts (Section \ref{subsec:annot-challenges}) and our web-based annotation tool (Section \ref{subsec:annot-tool}).

\begin{table}[!t]
\centering
\resizebox{0.44\textwidth}{!}
{%
  
 \begin{tabular}{ccc}
 \toprule 
            Script  & Source & Document Count \\
\midrule   
    Devanagari &  \textsc{PIH} & $193$          \\
    Nandinagari & \textsc{Bhoomi} & $2$          \\
    Telugu      & \textsc{Bhoomi} & $75$          \\
    Grantha     & \textsc{Bhoomi} & $238$          \\
 \bottomrule
 \end{tabular}
 }
\captionof{table}{Scripts in the \textsc{Indiscapes} dataset.}
\label{tab:dataset-script-stats} 
\end{table}

\subsection{Annotation Challenges}
\label{subsec:annot-challenges}

A variety of unique challenges exist in the context of annotating Indic manuscript layouts. The challenges arise from three major sources.

\noindent \textbf{Content:} The documents are written in a large variety of Indic languages. Some languages even exhibit intra-language script variations. A large pool of annotators familiar with the languages and scripts present in the corpus is required to ensure proper annotation of lines and character components.

\noindent \textbf{Layout:} Unlike some of the existing datasets, Indic manuscripts contain non-textual elements such as color pictures, tables and document decorations. These elements are frequently interspersed with text in non-standard layouts. In many cases, the  manuscripts contain one or more physical holes, designed for a thread-like material to pass through and bind the leaves together as a book. Such holes vary in terms of spatial location, count and hole diameter. When the holes are present in the middle of the document, they cause a break in the contiguity of lines. In some documents, the line contiguity is broken by a `virtual' hole-like gap, possibly intended for creation of the punched hole at a future time. In many cases, the separation between lines is extremely small. The handwritten nature of these documents and the surface material result in extremely uneven lines, necessitating meticulous and slow annotation. If multiple manuscript pages are present, the stacking order could be horizontal or vertical. Overall, the sheer variety in layout elements poses a significant challenge, not only for annotation, but also for automated layout parsing.

\noindent \textbf{Degradations:} Historical Indic manuscripts tend to be inherently fragile and prone to damage due to various sources -- wood-and-leaf-boring insects, humidity seepage, improper storage and handling etc. While some degradations cause the edges of the document to become frayed, others manifest as irregularly shaped perforations in the document interior. It may be important to identify such degradations before attempting lexically-focused tasks such as OCR or word-spotting. 

\begin{figure*}[hbt!]
\includegraphics[width=0.98\textwidth]{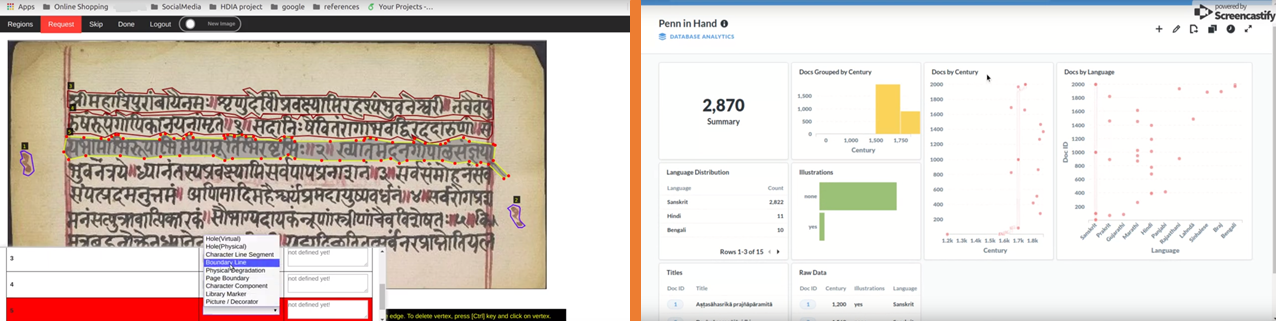}
\caption{Screenshots of our web-based annotator (left) and analytics dashboard (right).}
\label{fig:annot-tool-screenshot}
\end{figure*}

\subsection{Annotation Tool}
\label{subsec:annot-tool}

Keeping the aforementioned challenges in mind, we introduce a new browser-based annotation tool (see Figure \ref{fig:annot-tool-screenshot}). The tool is designed to operate both stand-alone and as a web-service. The web-service mode enables features such as distributed parallel sessions by registered annotators, dashboard-based live session monitoring and a wide variety of annotation-related analytics. On the front-end, a freehand region option is provided alongside the usual rectangle and polygon to enable maximum annotation flexibility. The web-service version also features a `Correction-mode' which enables annotators to correct existing annotations from previous annotators. Additionally, the tool has been designed to enable lexical (text) annotations in future.  

\section{Indic Manuscript Layout Parsing}
\label{sec:layout-parsing}

\begin{figure*}[hbt!]
\includegraphics[width=0.97\textwidth]{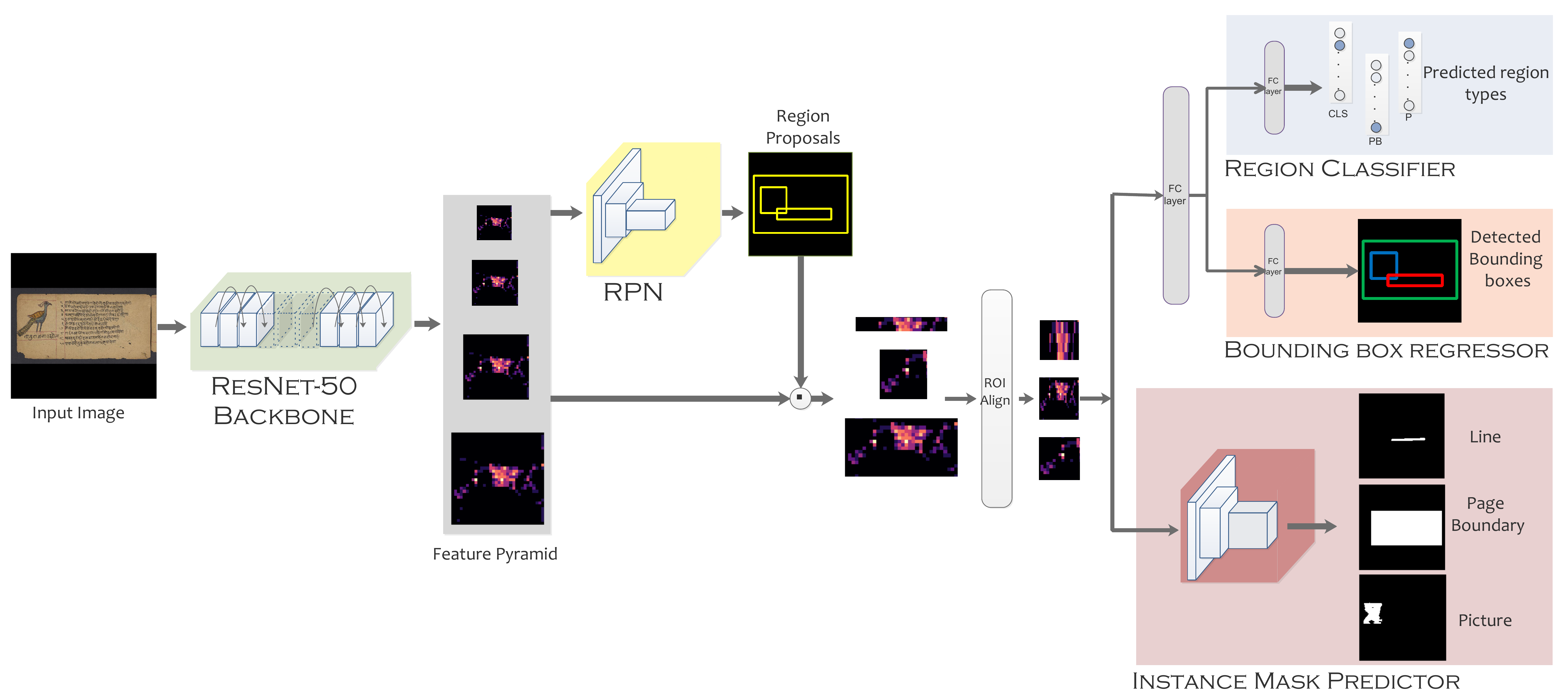}
\caption{The architecture adopted for Indic Manuscript Layout Parsing. Refer to Section \ref{sec:layout-parsing} for details.}
\label{fig:overview}
\end{figure*}


\begin{figure*}[phbt!]
\includegraphics[width=0.95\textwidth]{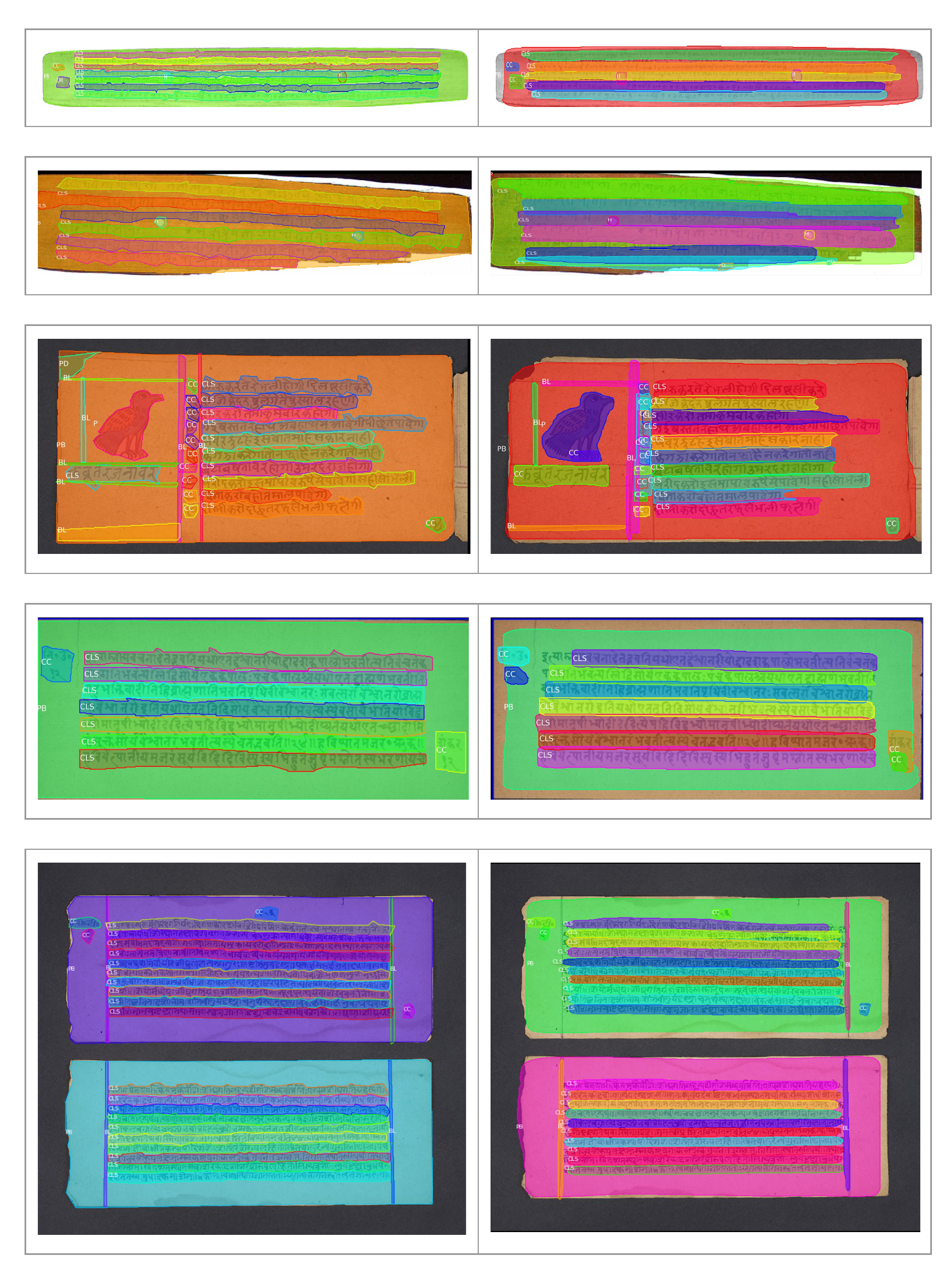}
\caption{Ground truth annotations (left) and predicted instance segmentations (right) for test set images. Note that we use colored shading only to visualize individual region instances and not to color-code region types. The region label abbreviations are shown alongside the regions. CLS : Character Line Segment, PB : Page Boundary, H : Hole, BL : Boundary Line, CC : Character Component, PD : Physical Degradation.}
\label{fig:results}
\end{figure*}

To succeed at layout parsing of manuscripts, we require a system which can accurately localize various types of regions (e.g. text lines, isolated character components, physical degradation, pictures, holes). More importantly, we require a system which can isolate individual \textit{instances} of each region (e.g. multiple text lines) in the manuscript image. Also, in our case, the annotation regions for manuscripts are not disjoint and can overlap (e.g. The annotation region for a text line can overlap with the annotation region of a hole (see Figure \ref{fig:results})). Therefore, we require a system which can accommodate such overlaps. To meet all of these requirements, we model our problem as one of semantic \textit{instance-level} segmentation and employ the Mask R-CNN~\cite{He2017MaskR} architecture which has proven to be very effective at the task of object-instance segmentation in photos. Next, we briefly describe the Mask R-CNN architecture and our modifications of the same. Subsequently, we provide details related to implementation (Section \ref{sec:implementation}), model training (Section \ref{sec:training}) and inference (Section \ref{sec:inference}).

\begin{table*}[!htbp]
\resizebox{\textwidth}{!}{%
\centering
 \centering 
 \begin{tabular}{lccccccccc}
 \toprule 
    & \multicolumn{9}{c}{Average IoU / Average Per pixel Accuracy}   \\
 Dataset $\downarrow$ &  \texttt{H} & \texttt{\raggedright{CLS}} & \texttt{PD} & \texttt{PB} & \texttt{CC} &  \texttt{P} & \texttt{D} & \texttt{LM} & \texttt{BL}  \\ 
 \midrule
 \textsc{PIH}            &  $-$ & $74.17/92.57$  & $-$ &   $86.90/96.37$      &  $52.84/74.85$       & $60.49/82.21$  & $5.23/6.17$ & $50.29/56.97$  & $29.45/43.14$  \\
 \textsc{Bhoomi}           &  $79.29/99.95$ & $29.07/43.67$  & $8.72/12.98$ & $91.09/99.22$ & $32.50/47.19$         & $-$  & $-$ & $38.25/49.98$  & $-$  \\ 
 Combined            &  $79.29/99.95$ & $57.77/74.79$  & $8.72/12.98$ & $88.47/97.44$ & $45.87/65.37$ & $60.49/82.21$  & $5.23/6.17$ & $42.93/52.70$  & $29.45/43.14$  \\ 
  
 \bottomrule
 \end{tabular}
 }
\captionof{table}{Class-wise average IoUs and per-pixel accuracies on the test set. Refer to Table \ref{tab:dataset-region-stats} for full names of abbreviated region types listed at top of the table.}
\label{tab:meas-1} 
\end{table*}

\begin{table}[!ht]
\centering
\resizebox{0.44\textwidth}{!}
{%
  
 \begin{tabular}{cccc}
 \toprule 
                & $AP_{50}$  &  $AP_{75}$ & $AP$  \\
\midrule   
    PIH        & 79.78        & 60.11        & 49.64                     \\
    Bhoomi     & 36.88        & 14.95         & 18.00                     \\
    Combined  
          &    64.76       & 44.30         & 38.57                     \\
 \bottomrule
 \end{tabular}
 }
\captionof{table}{AP at IoU thresholds $50$, $75$ and overall $AP$ averaged over IoU range for test set.}
\label{tab:meas-2} 
\end{table}

\subsection{Network Architecture}
\label{sec:layout-parsing-network}

The Mask-RCNN architecture contains three stages as described below (see Figure \ref{fig:overview}).  

\noindent \textbf{Backbone:} The first stage, referred to as the backbone, is used to extract features from the input image. It consists of a convolutional network combined with a feature-pyramid network \cite{DBLP:conf/cvpr/LinDGHHB17}, thereby enabling multi-scale features to be extracted. We use the first four blocks of ResNet-50 \cite{DBLP:conf/cvpr/HeZRS16} as the convolutional network. 

\noindent \textbf{Region Proposal Network (RPN):} This is a convolutional network which scans the pyramid feature map generated by the backbone network and generates rectangular regions commonly called `object proposals' which are likely to contain objects of interest. For each level of the feature pyramid and for each spatial location at a given level, a set of level-specific bounding boxes called anchors are generated. The anchors typically span a range of aspect ratios (e.g. $1:2, 1:1, 2:1$) for flexibility in detection. For each anchor, the RPN network predicts (i) the probability of an object being present (`objectness score') (ii) offset coordinates of a bounding box relative to location of the anchor. The generated bounding boxes are first filtered according to the `objectness score'. From boxes which survive the filtering, those that overlap with the underlying object above a certain threshold are chosen. After applying non-maximal suppression to remove overlapping boxes with relatively smaller objectness scores, the final set of boxes which remain are termed `object proposals' or Regions-of-Interest (RoI).

\noindent \textbf{Multi-Task Branch Networks:} The RoIs obtained from RPN are warped into fixed dimensions and overlaid on feature maps extracted from the backbone to obtain RoI-specific features. These features are fed to three parallel task sub-networks. The first sub-network maps these features to region labels (e.g. \texttt{Hole},\texttt{Character-Line-Segment}) while the second sub-network maps the RoI features to bounding boxes. The third sub-network is fully convolutional and maps the features to the pixel mask of the underlying region. Note that the ability of the architecture to predict masks independently for each RoI plays a  crucial role in obtaining instance segmentations. Another advantage is that it naturally addresses situations where annotations or predictions overlap. 

\subsection{Implementation Details}
\label{sec:implementation}

The dataset splits used for training, validation and test phases can be seen in Table \ref{table:document-counts}. All manuscript images are adaptively resized to ensure the width does not exceed $1024$ pixels. The images are padded with zeros such that the input to the deep network has spatial dimensions of $1024 \times 1024$. The ground truth region masks are initially subjected to a similar resizing procedure. Subsequently, they are downsized to $28 \times 28$ in order to match output dimensions of the mask sub-network.

\subsubsection{Training}
\label{sec:training}

The network is initialized with weights obtained from a Mask R-CNN trained on the MS-COCO~\cite{DBLP:journals/corr/LinMBHPRDZ14} dataset with a ResNet-50 backbone. We found that this results in faster convergence and stabler training compared to using weights from a Mask-RCNN trained on ImageNet~\cite{deng2009imagenet} or training from scratch. Within the RPN network, we use custom-designed anchors of $5$ different scales and with $3$ different aspect ratios. Specifically, we use the following aspect ratios -- 1:1,1:3,1:10 -- keeping in mind the typical spatial extents of the various region classes. We also limit the number of RoIs (`object proposals') to $512$. We use categorical cross entropy loss $\mathcal{L}_{RPN}$ for RPN classification network. Within the task branches, we use categorical cross entropy loss $\mathcal{L}_{r}$ for region classification branch, smooth L1 loss~\cite{ren2015faster} ($\mathcal{L}_{bb}$) for final bounding box prediction and per-pixel binary cross entropy loss $\mathcal{L}_{mask}$ for mask prediction. The total loss is a convex combination of these losses, i.e. $\mathcal{L} = \lambda_{RPN} \mathcal{L}_{RPN}  + \lambda_{r} \mathcal{L}_{r} + \lambda_{bb} \mathcal{L}_{bb} + \lambda_{mask} \mathcal{L}_{mask}$. The weighting factors ($\lambda$s) are set to $1$. However, to ensure priority for our task of interest namely mask prediction, we set $\lambda_{mask}=2$. For optimization, we use Stochastic Gradient Descent (SGD) optimizer with a gradient norm clipping value of $0.5$. The batch size, momentum and weight decay are set to $1$, $0.9$ and $10^{-3}$ respectively. Given the relatively smaller size of our manuscript dataset compared to the photo dataset (MS-COCO) used to originally train the base Mask R-CNN, we adopt a multi-stage training strategy. For the first stage ($30$ epochs), we train only the task branch sub-networks using a learning rate of $10^{-3}$ while freezing weights in the rest of the overall network. This ensures that the task branches are fine-tuned for the types of regions contained in manuscript images. For the second stage ($20$ epochs), we additionally train stage-4 and up of the backbone ResNet-50. This enables extraction of appropriate semantic features from manuscript images. The omission of the initial $3$ stages in the backbone for training is due to the fact that they provide generic, re-usable low-level features. To ensure priority coverage of hard-to-localize regions, we use focal loss~\cite{lin2017focal} for mask generation. For the final stage ($15$ epochs), we train the entire network using a learning rate of $10^{-4}$.

\subsubsection{Inference}
\label{sec:inference}

During inference, the images are rescaled and processed using the procedure described at the beginning of the subsection. The number of RoIs retained after non-maximal suppression (NMS) from the RPN is set to $1000$. From these, we choose the top $100$ region detections with objectness score exceeding $0.5$ and feed the corresponding RoIs to the mask branch sub-network for mask generation. It is important to note that this strategy is different from the parallel generation of outputs and use of the task sub-networks during training. The generated masks are then binarized using an empirically chosen threshold of $0.4$ and rescaled to their original size using bilinear interpolation. On these generated masks, NMS with a threshold value of $0.5$ is applied to obtain the final set of predicted masks.

\subsection{Evaluation}
\label{section:evaluation}

For quantitative evaluation, we compute Average Precision (AP) for a particular IoU threshold, a measure widely reported in instance segmentation literature ~\cite{DBLP:journals/corr/CordtsORREBFRS16,DBLP:journals/corr/LinMBHPRDZ14}. We specifically report $AP_{50}$ and $AP_{75}$, corresponding to AP at IoU thresholds 50 and 75 respectively~\cite{He2017MaskR}. In addition, we report an overall score by averaging  AP at different IoU thresholds ranging from $0.5$ to $0.95$ in steps of $0.05$. 

The AP measure characterizes performance at document level. To characterize performance for each region type, we report two additional measures~\cite{chen2017convolutional} -- average class-wise IoU (cwIoU) and average class-wise per-pixel accuracy (cwAcc). Consider a fixed test document $k$. Suppose there are $r_i$ regions of class $i$ and let ${IoU}_r$ denote the IoU score for one such region $r$, i.e. $1 \leqslant r \leqslant r_i$. 
The per-class IoU score for class $i$ and document $k$ is computed as ${cwIoU}^d_i = \frac{\sum_r {IoU}_r}{r_i}$. Suppose there are $N_i$ documents containing at least a single region of class $i$ in ground-truth. The overall per-class IoU score for class $i$ is computed as  ${cwIoU}_i = \frac{\sum_d {cwIoU}^d_i}{N_i}$. In a similar manner, we define class-wise pixel accuracy ${pwAcc}^d_i$ at document level and average it across all the documents containing class $i$, i.e. ${cwAcc}_i = \frac{\sum_d {pwAcc}^d_i}{N_i}$. Note that our approach for computing class-wise scores prevents documents with a relatively larger number of class instances from dominating the score and in this sense, differs from existing approaches~\cite{chen2017convolutional} 

\section{Results}
\label{section:results}

We report quantitative results using the measures described in Section \ref{section:evaluation}. Table \ref{tab:meas-1} reports Average Precision and Table \ref{tab:meas-2} reports class-wise average IOUs and per-pixel accuracies. Qualitative results can be viewed in Figure \ref{fig:results}. Despite the challenges posed by manuscripts, our model performs reasonably well across a variety of classes. As the qualitative results indicate, the model predicts accurate masks for almost all the regions. The results also indicate that our model handles overlap between \textit{Hole}s and \textit{Character line segment}s well. From  ablative experiments, we found that our choice of focal loss was crucial in obtaining accurate mask boundaries. Unlike traditional semantic segmentation which would have produced a single blob-like region for line segments, our instance-based approach isolates each text line separately. Additionally, the clear demarcation between \textit{Page-Boundary} and background indicates that our system identifies semantically relevant regions for downstream analysis. As the result at the bottom of Figure \ref{fig:results} shows, our system can even handle images with multiple pages, thus removing the need for any pre-processing related to isolation of individual pages.

From quantitative results, we observe that \textit{Holes}, \textit{Character line segments}, \textit{Page boundary} and \textit{Pictures} are parsed the best while \textit{Physical degradations} are difficult to parse due to the relatively small footprint and inconsistent patterns in degradations. The results show that performance for Penn in Hand (\textsc{PIH}) documents is better compared to \textsc{Bhoomi} manuscripts. We conjecture that the presence of closely spaced and unevenly written lines in latter is the cause. In our approach, two (or more) objects may share the same bounding box in terms of overlap and it is not possible to determine which box to choose during mask prediction. Consequently, an underlying line's boundary may either end up not being detected or the predicted mask might be poorly localized. However, this is not a systemic problem since our model achieves good performance even for very dense \textsc{Bhoomi} document line layouts.

\section{Conclusion}

Via this paper, we propose Indiscapes, the first dataset with layout annotations for historical Indic manuscripts. We believe that the availability of layout annotations will play a crucial role in reducing the overall complexity for OCR and other tasks such as word-spotting, style-and-content based retrieval. In the  long-term, we intend to expand the dataset, not only numerically but also in terms of layout, script and language diversity. As a significant contribution, we have also adapted a deep-network based instance segmentation framework custom modified for fully automatic layout parsing. Given the general nature of our framework, advances in instance segmentation approaches can be leveraged thereby improving performance over time. Our proposed web-based annotator system, although designed for Indic manuscripts, is flexible, and could be reused for similar manuscripts from Asian subcontinent. We intend to expand the capabilities of our annotator system in many useful ways. For instance, the layout estimated by our deep-network could be provided to annotators for correction, thus reducing annotation efforts. Finally, we plan to have our dataset, instance segmentation system and annotator system publicly available. This would enable large-scale data collection and automated analysis efforts for Indic as well as other historical Asian manuscripts. The repositories related to the systems presented in this paper and the Indiscapes dataset can be accessed at \url{https://ihdia.iiit.ac.in}.

\section*{Acknowledgment}
We would like to thank Dr. Sai Susarla for enabling access to the Bhoomi document collection. We also thank Poreddy Mourya Kumar Reddy, Gollapudi Sai Vamsi Krishna for their contributions related to dashboard and various annotators for their labelling efforts.
\bibliographystyle{IEEEtran}
\bibliography{references}

\begin{thebibliography}{10}
\providecommand{\url}[1]{#1}
\csname url@samestyle\endcsname
\providecommand{\newblock}{\relax}
\providecommand{\bibinfo}[2]{#2}
\providecommand{\BIBentrySTDinterwordspacing}{\spaceskip=0pt\relax}
\providecommand{\BIBentryALTinterwordstretchfactor}{4}
\providecommand{\BIBentryALTinterwordspacing}{\spaceskip=\fontdimen2\font plus
\BIBentryALTinterwordstretchfactor\fontdimen3\font minus
  \fontdimen4\font\relax}
\providecommand{\BIBforeignlanguage}[2]{{%
\expandafter\ifx\csname l@#1\endcsname\relax
\typeout{** WARNING: IEEEtran.bst: No hyphenation pattern has been}%
\typeout{** loaded for the language `#1'. Using the pattern for}%
\typeout{** the default language instead.}%
\else
\language=\csname l@#1\endcsname
\fi
#2}}
\providecommand{\BIBdecl}{\relax}
\BIBdecl

\bibitem{reul2017case}
C.~Reul, M.~Dittrich, and M.~Gruner, ``Case study of a highly automated layout
  analysis and ocr of an incunabulum:'der heiligen leben'(1488),'' in
  \emph{Proc. 2nd Intl. Conf. on Digital Access to Textual Cultural
  Heritage}.\hskip 1em plus 0.5em minus 0.4em\relax ACM, 2017, pp. 155--160.

\bibitem{springmann2017ocr}
U.~Springmann and A.~Luedeling, ``Ocr of historical printings with an
  application to building diachronic corpora: A case study using the ridges
  herbal corpus,'' \emph{Digital Humanities Quarterly}, no.~2, 2017.

\bibitem{simistira2016diva}
F.~Simistira, M.~Seuret, N.~Eichenberger, A.~Garz, M.~Liwicki, and R.~Ingold,
  ``Diva-hisdb: A precisely annotated large dataset of challenging medieval
  manuscripts,'' in \emph{ICFHR}.\hskip 1em plus 0.5em minus 0.4em\relax IEEE,
  2016, pp. 471--476.

\bibitem{PappoToledano2018AdoptiveTA}
A.~Pappo-Toledano, F.~Chen, G.~Latif, and L.~Alzubaidi, ``Adoptive thresholding
  and geometric features based physical layout analysis of scanned arabic
  books,'' \emph{2018 IEEE 2nd Intl. Workshop on Arabic and Derived Script
  Analysis and Recognition (ASAR)}, pp. 171--176, 2018.

\bibitem{kesiman2016amadi_lontarset}
M.~W.~A. Kesiman, J.-C. Burie, G.~N. M.~A. Wibawantara, I.~M.~G. Sunarya, and
  J.-M. Ogier, ``Amadi\_lontarset: The first handwritten balinese palm leaf
  manuscripts dataset,'' in \emph{ICFHR}.\hskip 1em plus 0.5em minus
  0.4em\relax IEEE, 2016, pp. 168--173.

\bibitem{chen2015page}
K.~Chen, M.~Seuret, M.~Liwicki, J.~Hennebert, and R.~Ingold, ``Page
  segmentation of historical document images with convolutional autoencoders,''
  in \emph{ICDAR}.\hskip 1em plus 0.5em minus 0.4em\relax IEEE, 2015, pp.
  1011--1015.

\bibitem{sahoo2016selective}
J.~Sahoo, ``A selective review of scholarly communications on palm leaf
  manuscripts,'' \emph{Library Philosophy and Practice (e-journal)}, 2016.

\bibitem{rachman2018palm}
Y.~B. Rachman, ``Palm leaf manuscripts from royal surakarta, indonesia:
  Deterioration phenomena and care practices,'' \emph{Intl. Journal for the
  Preservation of Library and Archival Material}, vol.~39, no.~4, pp. 235--247,
  2018.

\bibitem{kumar2009traditional}
D.~U. Kumar, G.~Sreekumar, and U.~Athvankar, ``Traditional writing system in
  southern india—palm leaf manuscripts,'' \emph{Design Thoughts}, vol.~9,
  2009.

\bibitem{valy2017new}
D.~Valy, M.~Verleysen, S.~Chhun, and J.-C. Burie, ``A new khmer palm leaf
  manuscript dataset for document analysis and recognition: Sleukrith set,'' in
  \emph{Proc. of the 4th Intl. Workshop on Historical Document Imaging and
  Processing}.\hskip 1em plus 0.5em minus 0.4em\relax ACM, 2017, pp. 1--6.

\bibitem{sanchez2014handwritten}
J.~A. S{\'a}nchez, V.~Bosch, V.~Romero, K.~Depuydt, and J.~De~Does,
  ``Handwritten text recognition for historical documents in the
  transcriptorium project,'' in \emph{Proc. of the First Intl. Conf. on Digital
  Access to Textual Cultural Heritage}.\hskip 1em plus 0.5em minus 0.4em\relax
  ACM, 2014, pp. 111--117.

\bibitem{rath2007word}
T.~M. Rath and R.~Manmatha, ``Word spotting for historical documents,''
  \emph{IJDAR}, vol.~9, no. 2-4, pp. 139--152, 2007.

\bibitem{kassis2017vmlhd}
M.~Kassis, A.~Abdalhaleem, A.~Droby, R.~Alaasam, and J.~El-Sana, ``Vml-hd: The
  historical arabic documents dataset for recognition systems,'' in \emph{1st
  Intl. Workshop on Arabic Script Analysis and Recognition}.\hskip 1em plus
  0.5em minus 0.4em\relax IEEE, 2017.

\bibitem{suryani2017handwritten}
M.~Suryani, E.~Paulus, S.~Hadi, U.~A. Darsa, and J.-C. Burie, ``The handwritten
  sundanese palm leaf manuscript dataset from 15th century,'' in
  \emph{ICDAR}.\hskip 1em plus 0.5em minus 0.4em\relax IEEE, 2017, pp.
  796--800.

\bibitem{clausner2017icdar2017}
C.~Clausner, A.~Antonacopoulos, T.~Derrick, and S.~Pletschacher, ``Icdar2017
  competition on recognition of early indian printed documents-reid2017,'' in
  \emph{ICDAR}, vol.~1.\hskip 1em plus 0.5em minus 0.4em\relax IEEE, 2017, pp.
  1411--1416.

\bibitem{Savitha_2018}
C.~K. Savitha and P.~J. Antony, ``Machine learning approaches for recognition
  of offline tulu handwritten scripts,'' \emph{Journal of Physics: Conference
  Series}, vol. 1142, p. 012005, nov 2018.

\bibitem{abeysinghe2018use}
A.~Abeysinghe and A.~Abeysinghe, ``Use of neural networks in archaeology:
  preservation of assamese manuscripts.''\hskip 1em plus 0.5em minus
  0.4em\relax International Seminar on Assamese Culture \& Heritage, 2018.

\bibitem{sastry20173d}
P.~N. Sastry, T.~V. Lakshmi, N.~K. Rao, and K.~RamaKrishnan, ``A 3d approach
  for palm leaf character recognition using histogram computation and distance
  profile features,'' in \emph{Proc. 5th Intl. Conf. on Frontiers in
  Intelligent Computing: Theory and Applications}.\hskip 1em plus 0.5em minus
  0.4em\relax Springer, 2017, pp. 387--395.

\bibitem{Panyam:2016:MPL:3043545.3064264}
N.~S. Panyam, V.~L. T.R., R.~Krishnan, and K.~R. N.V., ``Modeling of palm leaf
  character recognition system using transform based techniques,''
  \emph{Pattern Recogn. Lett.}, vol.~84, no.~C, Dec. 2016.

\bibitem{shi2004digital}
Z.~Shi, S.~Setlur, and V.~Govindaraju, ``Digital enhancement of palm leaf
  manuscript images using normalization techniques,'' in \emph{5th Intl. Conf.
  On Knowledge Based Computer Systems}, 2004, pp. 19--22.

\bibitem{malayalamPalm}
D.~Sudarsan, P.~Vijayakumar, S.~Biju, S.~Sanu, and S.~K. Shivadas,
  ``Digitalization of malayalam palmleaf manuscripts based on contrast-based
  adaptive binarization and convolutional neural networks,'' in \emph{Intl.
  Conf. on Wireless Communications, Signal Processing and Networking
  (WiSPNET)}, 2018.

\bibitem{wick2018fully}
C.~Wick and F.~Puppe, ``Fully convolutional neural networks for page
  segmentation of historical document images,'' in \emph{DAS}.\hskip 1em plus
  0.5em minus 0.4em\relax IEEE, 2018, pp. 287--292.

\bibitem{Wei2015}
H.~Wei, M.~Seuret, K.~Chen, A.~Fischer, M.~Liwicki, and R.~Ingold, ``Selecting
  autoencoder features for layout analysis of historical documents,'' in
  \emph{Proc. 3rd Intl. Workshop on Historical Document Imaging and
  Processing}, ser. HIP '15.\hskip 1em plus 0.5em minus 0.4em\relax ACM, 2015,
  pp. 55--62.

\bibitem{bukhari2012layout}
S.~S. Bukhari, T.~M. Breuel, A.~Asi, and J.~El-Sana, ``Layout analysis for
  arabic historical document images using machine learning,'' in \emph{ICFHR
  2012}.\hskip 1em plus 0.5em minus 0.4em\relax IEEE, 2012, pp. 639--644.

\bibitem{chen2017convolutional}
K.~Chen, M.~Seuret, J.~Hennebert, and R.~Ingold, ``Convolutional neural
  networks for page segmentation of historical document images,'' in
  \emph{ICDAR}, vol.~1.\hskip 1em plus 0.5em minus 0.4em\relax IEEE, 2017, pp.
  965--970.

\bibitem{barakat2018text}
B.~Barakat, A.~Droby, M.~Kassis, and J.~El-Sana, ``Text line segmentation for
  challenging handwritten document images using fully convolutional network,''
  in \emph{ICFHR}.\hskip 1em plus 0.5em minus 0.4em\relax IEEE, 2018, pp.
  374--379.

\bibitem{DBLP:conf/icfhr/KesimanVBPSHVCO18}
M.~W.~A. Kesiman, D.~Valy, J.~Burie, E.~Paulus, M.~Suryani, S.~Hadi,
  M.~Verleysen, S.~Chhun, and J.~Ogier, ``{ICFHR} 2018 competition on document
  image analysis tasks for southeast asian palm leaf manuscripts,'' in
  \emph{ICFHR}, 2018, pp. 483--488.

\bibitem{DBLP:conf/icdar/2017hip}
\emph{Proc. 4th Intl. Workshop on Historical Document Imaging and Processing,
  Kyoto, Japan, November 10-11, 2017}.\hskip 1em plus 0.5em minus 0.4em\relax
  {ACM}, 2017.

\bibitem{DBLP:conf/icdar/2015hip}
\emph{Proc. 3rd Intl. Wksp on Historical Document Imaging and Processing,
  HIP@ICDAR 2015}.\hskip 1em plus 0.5em minus 0.4em\relax {ACM}, 2015.

\bibitem{Sabeenian:2017:CHT:3133793.3133804}
R.~S. Sabeenian, M.~E. Paramasivam, P.~M. Dinesh, R.~Adarsh, and G.~R. Kumar,
  ``Classification of handwritten tamil characters in palm leaf manuscripts
  using svm based smart zoning strategies,'' in \emph{ICBIP}.\hskip 1em plus
  0.5em minus 0.4em\relax ACM, 2017.

\bibitem{kesiman2018benchmarking}
M.~W.~A. Kesiman, D.~Valy, J.-C. Burie, E.~Paulus, M.~Suryani, S.~Hadi,
  M.~Verleysen, S.~Chhun, and J.-M. Ogier, ``Benchmarking of document image
  analysis tasks for palm leaf manuscripts from southeast asia,'' \emph{Journal
  of Imaging}, vol.~4, no.~2, p.~43, 2018.

\bibitem{Valy2018Khmer}
D.~Valy, M.~Verleysen, S.~Chhun, and J.-C. Burie, ``Character and text
  recognition of khmer historical palm leaf manuscripts,'' in \emph{ICFHR}, 08
  2018, pp. 13--18.

\bibitem{Paulus2018}
E.~Paulus, M.~Suryani, and S.~Hadi, ``Improved line segmentation framework for
  sundanese old manuscripts,'' \emph{Journal of Physics: Conference Series},
  vol. 978, p. 012001, mar 2018.

\bibitem{doermann2010gedi}
D.~Doermann, E.~Zotkina, and H.~Li, ``{GEDI}-a groundtruthing environment for
  document images,'' in \emph{Ninth IAPR Intl. Workshop on Document Analysis
  Systems}, 2010.

\bibitem{garz2016creating}
A.~Garz, M.~Seuret, F.~Simistira, A.~Fischer, and R.~Ingold, ``Creating ground
  truth for historical manuscripts with document graphs and scribbling
  interaction,'' in \emph{DAS}.\hskip 1em plus 0.5em minus 0.4em\relax IEEE,
  2016, pp. 126--131.

\bibitem{clausner2011aletheia}
C.~Clausner, S.~Pletschacher, and A.~Antonacopoulos, ``Aletheia-an advanced
  document layout and text ground-truthing system for production
  environments,'' in \emph{ICDAR}.\hskip 1em plus 0.5em minus 0.4em\relax IEEE,
  2011, pp. 48--52.

\bibitem{webaletheia}
\BIBentryALTinterwordspacing
``Web aletheia.'' [Online]. Available:
  \url{https://github.com/PRImA-Research-Lab/prima-gwt-lib}
\BIBentrySTDinterwordspacing

\bibitem{wursch2016divaservices}
M.~W{\"u}rsch, R.~Ingold, and M.~Liwicki, ``Divaservices—a restful web
  service for document image analysis methods,'' \emph{Digital Scholarship in
  the Humanities}, vol.~32, no.~1, pp. i150--i156, 2016.

\bibitem{gatos2014ground}
B.~Gatos, G.~Louloudis, T.~Causer, K.~Grint, V.~Romero, J.~A. S{\'a}nchez,
  A.~H. Toselli, and E.~Vidal, ``Ground-truth production in the transcriptorium
  project,'' in \emph{DAS}.\hskip 1em plus 0.5em minus 0.4em\relax IEEE, 2014,
  pp. 237--241.

\bibitem{penninhand}
``Penn in hand: Selected manuscripts,''
  \url{http://dla.library.upenn.edu/dla/medren/search.html?fq=collection_facet:"Indic
  Manuscripts"}.

\bibitem{He2017MaskR}
K.~He, G.~Gkioxari, P.~Doll{\'a}r, and R.~B. Girshick, ``Mask r-cnn,''
  \emph{ICCV}, pp. 2980--2988, 2017.

\bibitem{DBLP:conf/cvpr/LinDGHHB17}
T.~Lin, P.~Doll{\'{a}}r, R.~B. Girshick, K.~He, B.~Hariharan, and S.~J.
  Belongie, ``Feature pyramid networks for object detection,'' in \emph{CVPR},
  2017, pp. 936--944.

\bibitem{DBLP:conf/cvpr/HeZRS16}
K.~He, X.~Zhang, S.~Ren, and J.~Sun, ``Deep residual learning for image
  recognition,'' in \emph{CVPR}, 2016, pp. 770--778.

\bibitem{DBLP:journals/corr/LinMBHPRDZ14}
\BIBentryALTinterwordspacing
T.~Lin, M.~Maire, S.~J. Belongie, L.~D. Bourdev, R.~B. Girshick, J.~Hays,
  P.~Perona, D.~Ramanan, P.~Doll{\'{a}}r, and C.~L. Zitnick, ``Microsoft
  {COCO:} common objects in context,'' \emph{CoRR}, vol. abs/1405.0312, 2014.
  [Online]. Available: \url{http://arxiv.org/abs/1405.0312}
\BIBentrySTDinterwordspacing

\bibitem{deng2009imagenet}
J.~Deng, W.~Dong, R.~Socher, L.-J. Li, K.~Li, and L.~Fei-Fei, ``Imagenet: A
  large-scale hierarchical image database,'' in \emph{CVPR}.\hskip 1em plus
  0.5em minus 0.4em\relax IEEE, 2009, pp. 248--255.

\bibitem{ren2015faster}
S.~Ren, K.~He, R.~Girshick, and J.~Sun, ``Faster r-cnn: Towards real-time
  object detection with region proposal networks,'' in \emph{NIPS}, 2015, pp.
  91--99.

\bibitem{lin2017focal}
T.-Y. Lin, P.~Goyal, R.~Girshick, K.~He, and P.~Doll{\'a}r, ``Focal loss for
  dense object detection,'' in \emph{ICCV}, 2017, pp. 2980--2988.

\bibitem{DBLP:journals/corr/CordtsORREBFRS16}
\BIBentryALTinterwordspacing
M.~Cordts, M.~Omran, S.~Ramos, T.~Rehfeld, M.~Enzweiler, R.~Benenson,
  U.~Franke, S.~Roth, and B.~Schiele, ``The cityscapes dataset for semantic
  urban scene understanding,'' \emph{CoRR}, vol. abs/1604.01685, 2016.
  [Online]. Available: \url{http://arxiv.org/abs/1604.01685}
\BIBentrySTDinterwordspacing

\end{thebibliography}
\end{document}